\documentclass{article}
\usepackage{spconf,amsmath,graphicx,hyperref}
\usepackage{amsmath, amssymb}
\usepackage{cite}

\title{StyMam: A Mamba-Based Generator for Artistic Style Transfer}
\name{
\parbox{\linewidth}{\centering
Zhou Hong$^{1,2,3,4}$, 
Ning Dong\textsuperscript{1,}\sthanks{Thanks to suqian sci\&tech k202349 for funding},
Yicheng Di\textsuperscript{2,}\sthanks{Corresponding author: diyicheng1@stu.jiangnan.edu.cn},
Rongsheng Hu$^{2}$,
Xiaolong Xu$^{5}$,
Yihua Shao$^{6}$, \\[0.3em]
Run Ling$^{7}$,
Yun Wang$^{7}$,
Juqin Wang$^{8}$,
Zhanjie Zhang$^{7}$,
Ao Ma$^{7}$
}
}
\address{
    $^{1}$School of Information Engineering, Suqian University, Suqian, China \\
    $^{2}$School of Artificial Intelligence and Computer Science, Jiangnan University, Wuxi, China\\
    $^{3}$Jiangsu Key Laboratory of Intelligent Production and Digital Technology for Brewing Raw Materials \\
    for Chinese Liquor, Suqian University, Suqian, China\\
    $^{4}$Jiangsu Province Engineering Research Center of Smart Poultry Farming and Intelligent Equipment\\
    , Suqian University, Suqian, China\\
    $^{5}$School of Computer Science, Nanjing University of Posts and Telecommunications, Nanjing, China\\
    $^{6}$ Institute of Automation, Chinese Academy of Sciences, Beijing, China \\
    $^{7}$JD.com, Beijing, China \\
    $^{8}$School of Internet of Things, Wuxi University of Technology, Wuxi, China\\
}
\begin{document}
\maketitle
\begin{abstract}
Image style transfer aims to integrate the visual patterns of a specific artistic style into a content image while preserving its content structure. Existing methods mainly rely on the generative adversarial network (GAN) or stable diffusion (SD). GAN-based approaches using CNNs or Transformers struggle to jointly capture local and global dependencies, leading to artifacts and disharmonious patterns. SD-based methods reduce such issues but often fail to preserve content structures and suffer from slow inference. To address these issues, we revisit GAN and propose a mamba-based generator, termed as StyMam, to produce high-quality stylized images without introducing artifacts and disharmonious patterns. Specifically, we introduce a mamba-based generator with a residual dual-path strip scanning mechanism and a channel-reweighted spatial attention module. The former efficiently captures local texture features, while the latter models global dependencies. Finally, extensive qualitative and quantitative experiments demonstrate that the proposed method outperforms state-of-the-art algorithms in both quality and speed.
\end{abstract}


\begin{figure*}[htb]  
\begin{minipage}[b]{1.0\linewidth}
  \centering
  \centerline{\includegraphics[width=12.8cm]{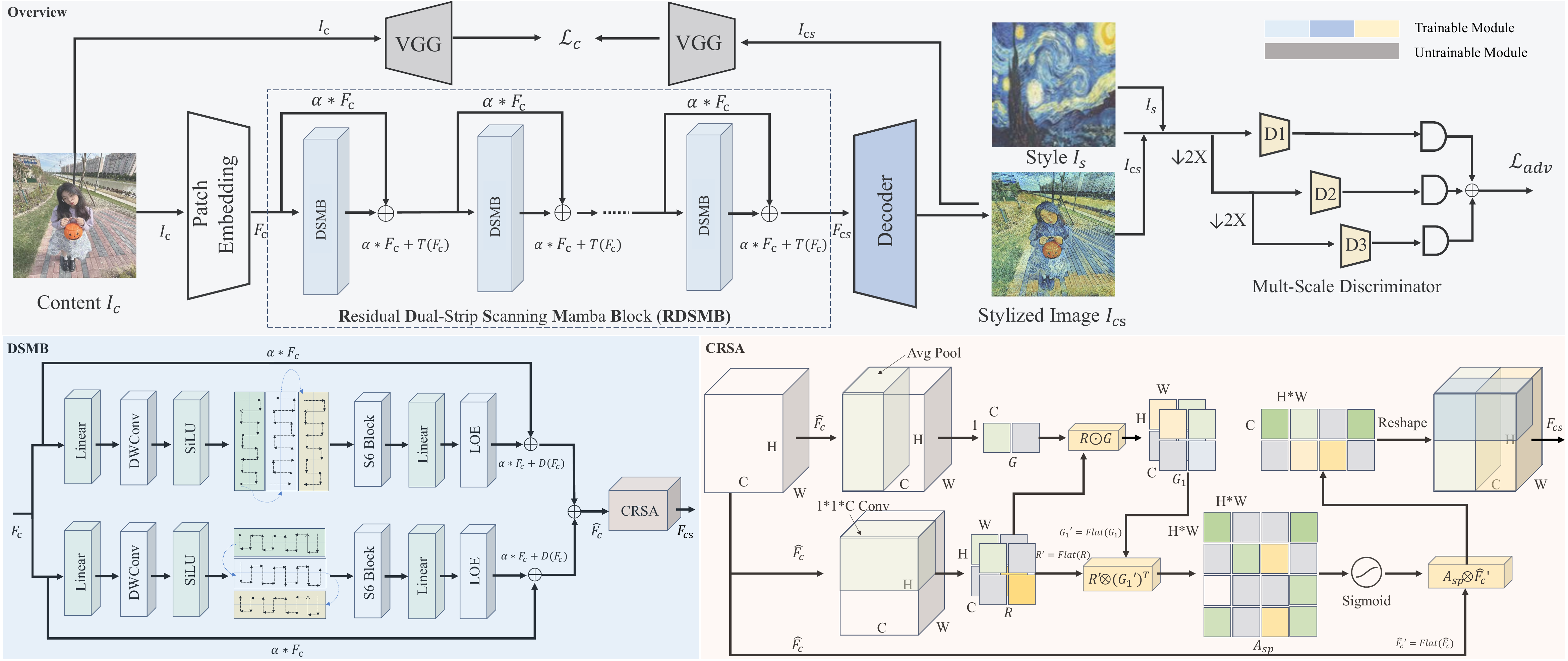}}
  \centerline{}\medskip
\end{minipage}
\vspace{-1.2cm}
\caption{The overview network of StyMam.}
\vspace{-0.5cm}
\label{fig:overvew}
\end{figure*}
\section{Introduction}
\label{sec:intro}

Image style transfer aims to apply specific artistic styles to content images while preserving their structure. Existing research efforts can be broadly categorized into generative adversarial network-based~\cite{park2020contrastive,zhu2017unpaired,zheng2021spatially,kim2023unpaired,sanakoyeu2018style} and stable diffusion-based~\cite{nichol2021glide,wang2022aesust,zhang2024towards} approaches. 

Specifically, some methods~\cite{park2020contrastive,zhu2017unpaired,zheng2021spatially,kim2023unpaired,sanakoyeu2018style} leverage generative adversarial network (GAN) on unpaired images to build the mapping between source and target domains. For example, CycleGAN~\cite{zhu2017unpaired} employs CNN-based networks along with a cycle consistency loss to enable bidirectional mapping between source and target domains. TransGAN~\cite{jiang2021transgan} incorporates attention mechanisms to capture long-range dependencies. However, both methods struggle to simultaneously capture local and global dependencies. This limitation often gives rise to artifacts and incongruent patterns and impacts the ability to generate high-quality stylized images.



Stable diffusion (SD)-based approaches \cite{nichol2021glide,wang2022aesust} utilize large-scale parameters to synthesize realistic images with fewer artifacts. For instance, Artbank~\cite{zhang2024artbank} proposes a global prompt space, a learnable parameter matrix, to store and condition the pre-trained diffusion model for generating artistic stylized images. LSAST~\cite{zhang2024towards} introduces a step-aware and a layer-aware prompt space with learnable prompts that adaptively adjust the content structures and style patterns of input images. While these SD-based approaches can produce highly realistic stylized images, they often struggle to preserve the content structure of the input image, leading to undesirable content details. Moreover, these methods are plagued by slow inference speeds, which significantly impact their efficiency.

To address the above challenges, we revisit GAN and propose a mamba-based generator for image style transfer, termed as StyMam. 
Mamba is a state-space model that captures global and local features in image processing~\cite{gu2023mamba,hong2025unpaired}, but its performance significantly depends on the scanning mechanism.
Although various scanning strategies~\cite{huang2024localmamba,li2025mair} have been proposed, they are relatively computationally complex for style transfer tasks. We argue that artistic style transfer focuses on learning the mapping from the source domain to the target domain, and introducing excessive scanning operations is unnecessary. As a result, we propose a mamba-based generator equipped with a dual-path strip scanning mechanism and a channel-reweighted spatial attention module. The former effectively captures local texture details, while the latter models global dependencies. Comprehensive qualitative and quantitative experiments show that the proposed method surpasses state-of-the-art algorithms in terms of both image quality and inference speed.
\section{METHODOLOGY}\nopagebreak
\label{sec:format}
\subsection{Overview}
\label{ssec:subhead}
The overall architecture of StyMam is illustrated in Fig.~\ref{fig:overvew}. StyMam consists of a mamba-based generator, a decoder, and a multi-scale discriminator. The generator incorporates a residual dual-strip scanning mamba block (RDSMB) and a channel-reweighted spatial attention (CRSA) module, which are described in Sec.~\ref{ssec:rdsmb} and~\ref{ssec:CRSA} respectively.
\begin{figure*}[htbp]  
  \centering
  \includegraphics[width=11.5cm]{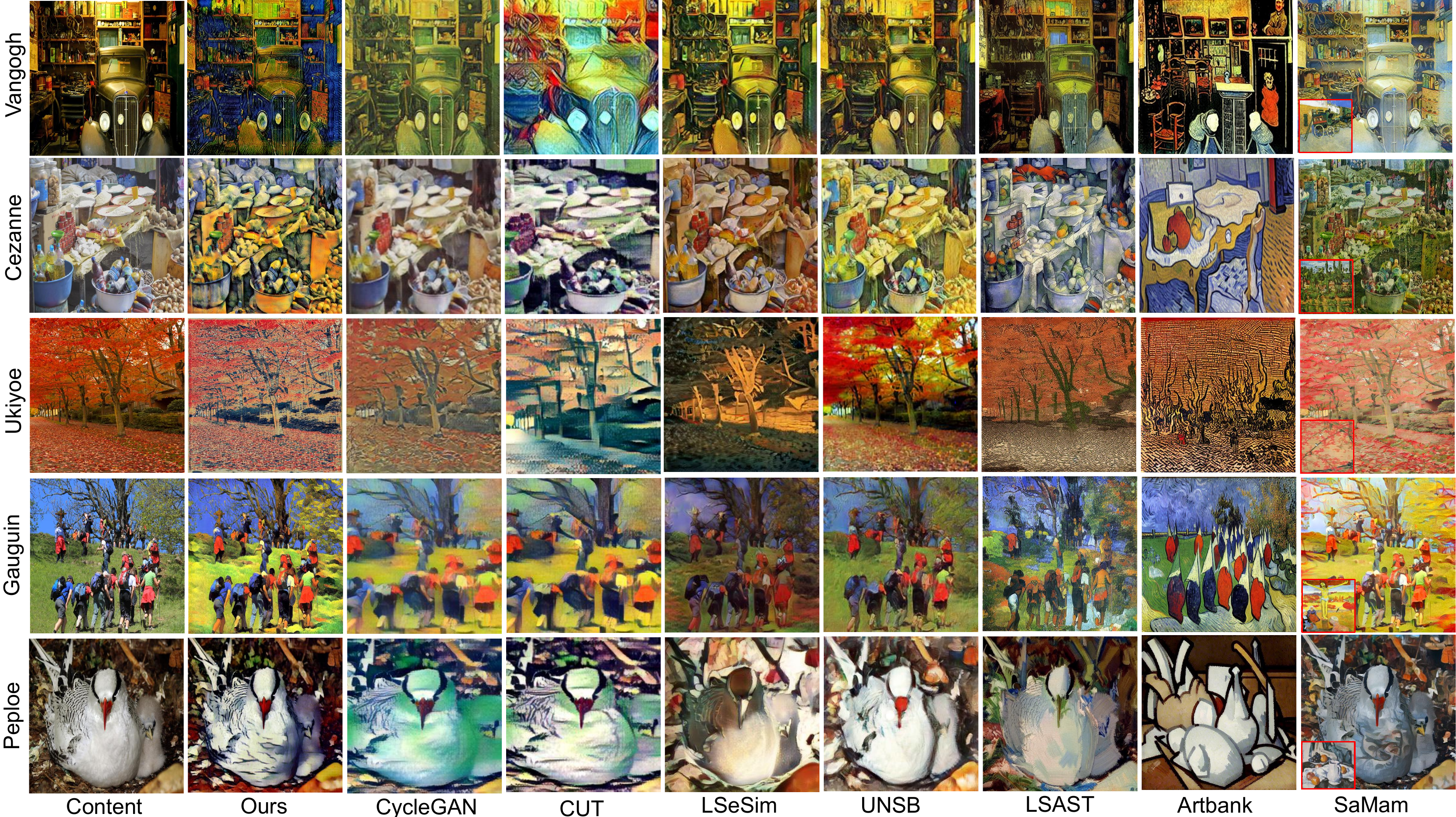}
\vspace{-0.3cm}
  \caption{Qualitative comparisons with SOTA diffusion-based and GAN-based style transfer methods.}
  \label{fig:res}
  \vspace{-0.5cm}
\end{figure*}
\subsection{RDSMB}
\label{ssec:rdsmb}
As illustrated in Fig.~\ref{fig:overvew}, the RDSMB is constructed by stacking multiple dual-strip scanning mamba block (DSMB) and residual connections. The DSMB consists of two parallel
branches, where each branch follows the computation flow
 of Linear → DWConv → SiLU → SS2D → Linear → LOE same as SaMam~\cite{liu2025samam}. 
A residual connection is introduced in each branch, scaled by a factor $\alpha$, to enhance training stability.

The mamba is based on a structured state space model (SSM), which can efficiently capture long-range dependencies while maintaining linear complexity~\cite{gu2023mamba}. 
Specially, given a content image $I_c$, we first embed it into a sequence (i.e., $\{\mathbf{F}_{c,t}\}_{t=1}^{D} = Patch(I_c)$) and then the state update and output can be formulated as:
\begin{equation}
\begin{aligned}
\label{eq:mamba}
\mathbf{h}_t &= A \mathbf{h}_{t-1} + B \mathbf{F}_{c,t}, \\
\hat{\mathbf{F}}_{c,t} &= C \mathbf{h}_t + D \mathbf{F}_{c,t}, \quad t = 1, \dots, D,
\end{aligned}
\end{equation}
where \(\mathbf{h}_t \in \mathbb{R}^{N}\) denotes the hidden state, \(\hat{\mathbf{F}}_{c,t} \in \mathbb{R}^{C}\) denotes the output at time step \(t\), and \(A, B, C, D\) are learnable matrices.
We fed \(\{\hat{\mathbf{F}}_{c,t}\}_{t=D}^{D}\) into the CRSA module which is describled in Sec.~\ref{ssec:CRSA}.

In fact, we employ the S6 block~\cite{liu2024vmamba} as the mamba module, which extends the standard SSM with diagonalization and bilinear parameterization. Also, image features are inherently 2D and must be scanned into 1D sequences before being processed by mamba. Equation~\ref{eq:mamba} only describes a single-path scanning. To efficiently capture local texture features, we adopt a dual-path strip scanning mechanism. We avoid complex scanning strategies, as style transfer primarily focuses on mapping a content image to its stylized counterpart. In contrast, tasks such as image restoration~\cite{li2025mair,hu2024zigma,huang2024localmamba} prioritize content reconstruction and require more detailed image information, which relies on more extensive scanning strategies.

As shown in Fig.~\ref{fig:overvew}, our proposed dual-path stip scanning method divides the feature map into several non-overlapping strips. Within and between strips, we use zigzag scanning~\cite{hu2024zigma} to preserve locality and spatial continuity. Unlike the MaIR~\cite{li2025mair}, we adopt a dual-path scanning strategy, processing the feature maps of each scanning method sequentially before inputting them into the CRSA (Sec. \ref{ssec:CRSA}). Specifically, the two different strip scanning orders are horizontal and vertical strip divisions, which illustrate the scanning order within each strip in Fig.~\ref{fig:overvew}. By employing the dual-path strip scanning mechanism, StyMam can more effectively capture local texture features.

\subsection{CRSA}
\label{ssec:CRSA}
To efficiently model global dependencies, we revisit the spatial attention and propose a channel reweighted spatial attention module.
As depicted in Fig.~\ref{fig:overvew}, given an input feature map 
\(\hat{F}_c \in \mathbb{R}^{H \times W \times C}
\), 
the proposed CRSA module first applies global average pooling to obtain a channel attention map 
\(G \in \mathbb{R}^{1 \times 1 \times C}\), 
and then performs a \(1 \times 1\) convolution to generate a feature map 
\(R \in \mathbb{R}^{H \times W \times C}\). 
For clarity, we denote the flattening operation \((\cdot)'\) that reshapes a tensor into a $2D$ matrix. 
The subsequent operations are as follows.

\textbf{Step 1.}  
The feature map feature $R$ is reweighted by the channel attention map $G$:
\begin{equation}
G_1 = R \odot G, \quad G_1 \in \mathbb{R}^{H \times W \times C},
\end{equation}
where \(\odot\) denotes element-wise multiplication. This step injects global channel priors into spatial features.  

\textbf{Step 2.}  
The flattened representations are used to compute a spatial attention map:
\begin{equation}
\mathbf{A}_{sp} = \mathbf{R}' \otimes (\mathbf{G}_1')^{\top}, \quad \mathbf{A}_{sp} \in \mathbb{R}^{HW \times HW},
\end{equation}
This operation explicitly models spatial dependencies among pixels.  

\textbf{Step 3.}  
Finally, the attention map is applied to reweight the original features:
\begin{equation}
\mathbf{F}'_{cs} = \text{Reshape}(\mathbf{A}_{sp} \otimes (\hat{\mathbf{F}}'_c)), 
\quad \mathbf{F}'_{cs} \in \mathbb{R}^{H \times W \times C},
\end{equation}
where \(\text{Reshape}(\cdot)\) restores the spatial dimension. This step integrates channel-reweighted spatial correlations with the original feature map, yielding enhanced feature representation. In summary, the proposed CRSA module integrates global channel priors with spatial dependency modeling, enabling adaptive emphasis on informative channels while capturing long-range spatial correlations, thereby enhancing content preservation.
\subsection{Loss Function}
\label{ssec:subhead}
\textbf{Content Loss:}  
To enforce structural consistency between the content image $I_c$ and the stylized result $I_{cs}$, we adopt a perceptual loss~\cite{huang2017arbitrary} based on VGG features:
\begin{equation}
\mathcal{L}_c = \sum_{i=1}^{4} \left\| \phi_i(I_c) - \phi_i(I_{cs}) \right\|_2^2,
\label{eq:content_loss}
\end{equation}
where $\phi_i(\cdot)$ denotes the feature representation from the $i$-th VGG layer.

\textbf{Adversarial Loss:}  
To enhance realism and suppress artifacts, we adopt adversarial learning with a multi-scale discriminator, which distinguishes real style images $I_s$ from generated stylized images $I_{cs}$. The adversarial loss is defined as:
\begin{multline}
\mathcal{L}_{adv} = \frac{1}{M} \sum_{m=1}^{M} \Big(
\mathbb{E}_{I_s \sim p_{data}} \big[\log D_m(I_s)\big] \\
+ \mathbb{E}_{I_c \sim p_c} \big[\log (1 - D_m(G(I_c)))\big]
\Big),
\label{eq:adv_loss}
\end{multline}
where $M$ denotes the number of scales, $D_m(\cdot)$ the $m$-th represents discriminator, 
$G(\cdot)$ is the generator, and $I_s$  stands for the style image. 

The objective function of StyMam is defined as a weighted sum of content and adversarial losses:
\begin{equation}
\mathcal{L}_{total} = \lambda_c \mathcal{L}_c + \lambda_{adv} \mathcal{L}_{adv},
\label{eq:total_loss}
\end{equation}
where $\lambda_c$ and $\lambda_{adv}$ are trade-off parameters, set to 1.0 and 5.0 respectively.

			
\begin{table*}[htb]
	\caption{Quantitative comparisons. The lower the FID score, the better the image quality. * denotes the average user preference.}
	\centering
	\setlength{\tabcolsep}{0.1cm}
	\begin{center}		
		\begin{tabular}{c|c|cccccccc}
			\hline
			\footnotesize Metrics & Datasets & Ours & CycleGAN & CUT & LSeSim & UNSB & LSAST & ArtBank & SaMam \\
			\hline
			\footnotesize  & Van Gogh &\textbf{93.80}&112.04&95.61&105.21&104.35&94.17&121.62&108.54 \\
			\footnotesize  & Ukiyoe   &\textbf{91.73}&140.78&115.44&130.77&110.29&92.03&111.50&135.07 \\
			\footnotesize \normalsize{FID}$\downarrow$ & Cezanne  &\textbf{129.02}&151.14&141.34&141.33&135.79&130.28&143.90&146.36 \\
			\footnotesize & Gauguin  &\textbf{143.07}&160.00&165.69&172.18&158.93&143.40&155.27&159.48 \\
			\footnotesize & Peploe   &\textbf{157.19}&167.77&164.39&192.88&165.21&158.44&182.51& 166.55\\
			\hline
			\footnotesize Time/sec $\downarrow$ & - &\textbf{0.0295}&0.0312&0.0312&0.0365&0.325&4.1325&3.7547& 0.041 \\
			\footnotesize Preference$\uparrow$ & - &\textbf{0.620}*&\textbf{0.702}/0.298&\textbf{0.713}/0.287&\textbf{0.694}/0.306&\textbf{0.606}/0.394&\textbf{0.517}/0.483&\textbf{0.541}/0.459&\textbf{0.568}/0.432\\
			\footnotesize Deception$\uparrow$ & - &\textbf{0.732}&{0.579}&{0.513}&{0.539}&{0.587}&{0.722}&{0.615}& {0.594}\\
			\hline 
		\end{tabular}
	\end{center}
	\label{FID}
    \vspace{-0.5cm}
\end{table*}
\begin{figure}[htb]  
  \centering
  \includegraphics[width=\linewidth]{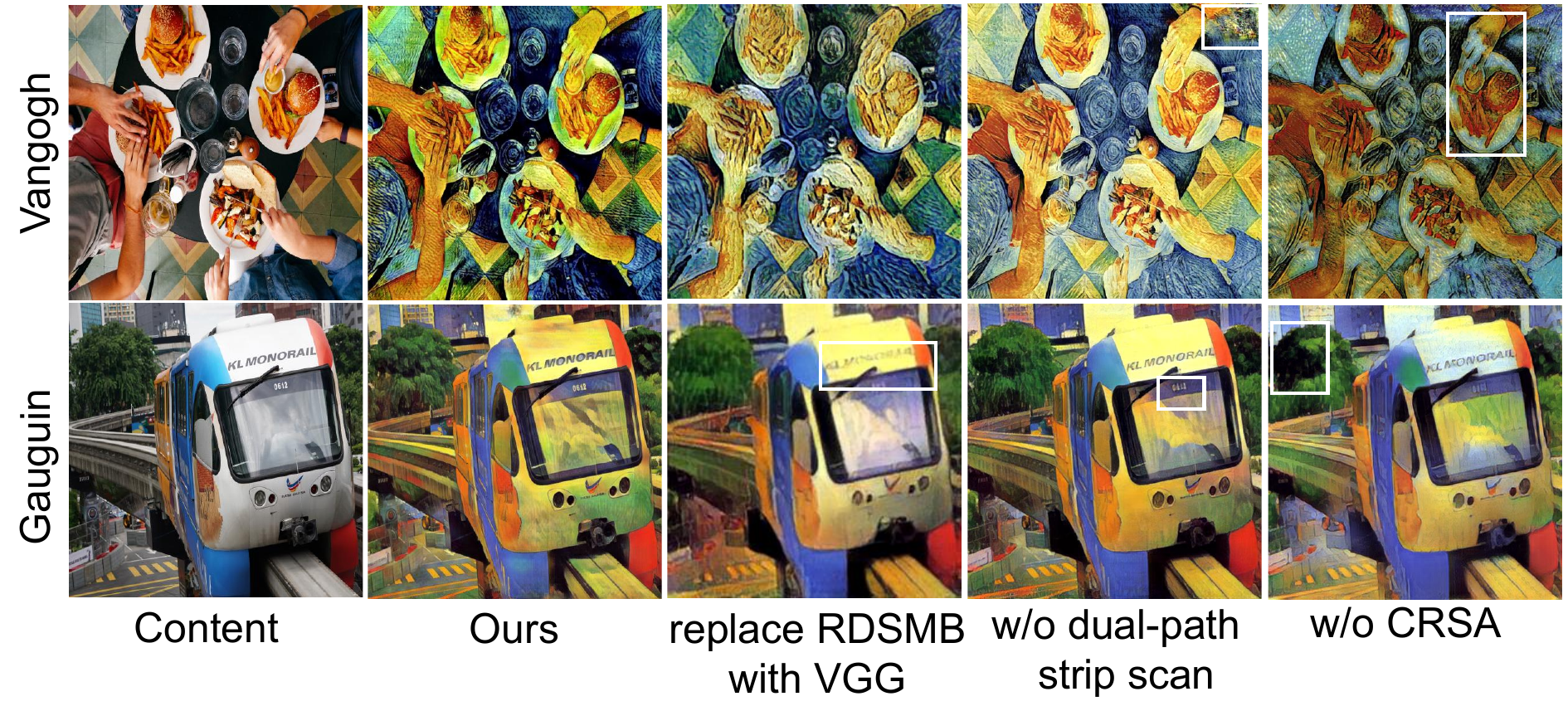}
  \vspace{-0.8cm}
  \caption{Ablation studies of our StyMam.}
  \label{fig:ab}
\end{figure}
\section{Experiments}
We implement our approach using two NVIDIA RTX 3090 GPUs with PyTorch. During training, we collect seven collections of artworks (e.g., Van Gogh, Cezanne, Peploe, Ukiyoe and Gauguin) from the Wikiart~\cite{zhang2025dyartbank} as style images and use content images from CycleGAN. All style and content images are resized to $256 \times 256$ pixels. For inference, we randomly select content images from the DIV2K~\cite{zhang2025dyartbank} and resize them to $512 \times 512$ pixels as the initial input content image.
\subsection{Qualitive comparisons}
As shown in Fig.\ref{fig:res}, we compare our method with state-of-the-art style transfer approaches, including CycleGAN~\cite{zhu2017unpaired}, LSeSim~\cite{zheng2021spatially}, UNSB~\cite{kim2023unpaired}, LSAST~\cite{zhang2024towards}, Artbank~\cite{zhang2024artbank}, and SaMam~\cite{liu2025samam}. Among them, CycleGAN, LSeSim, UNSB, and SaMam require retraining a well-designed forward network on a small dataset, which often fails to produce high-quality style transfer results and tends to introduce noticeable artifacts and inconsistent patterns. In contrast, Artbank and LSAST exploit implicit style prompts to leverage the rich style priors of pre-trained stable diffusion, which allows them to generate high-quality images but fails to preserve the content structure of the input images. By comparison, our method not only generates high-quality images but also maintains the structural integrity of the content.
\subsection{Quantitative comparisons}
\textbf{Frechet Inception Distance (FID).} We evaluate our method and other state-of-the-art artistic style transfer methods by generating 1,000 artistic images for each method and calculating the FID score between the generated images and real artistic images. As shown in Table~\ref{FID}, our method achieves a better FID score, indicating that the artistic stylized images generated by our method are closer to real artistic images.

\textbf{Preference Score.} We conducted an A/B test user study to compare the stylization effects between our method and other state-of-the-art methods. We randomly selected 100 content images, and for each image, we generated two stylized images: one from our method and the other from a randomly selected state-of-the-art method. We collected 5,000 votes from 50 subjects. As shown in Table~\ref{FID}, our method received the highest preference score.

\textbf{Deception Score.} The deception score evaluates whether the stylized images are perceived as human-created. A higher deception score means that the generated stylized images are more likely to be regarded as human-created artworks. We calculated the deception score based on 20 content images and 50 subjects. Our method achieved a score of 0.732, which is closer to the human-created artistic images, as shown in Table~\ref{FID}.

\textbf{Time Information.} The “Time/sec” row in Table~\ref{FID} shows the inference time of our method on $512 \times 512$ pixel images for comparison.
\subsection{Ablation Study}
To further verify the generator of the StyMam, we split it into dual-path stip scanning mechanism and CRSA module. As shown in Fig.~\ref{fig:ab}, the stylized images keeps less texture without dual-path strip scan. If removing CRSA, our proposed method can't preserve some content details and show some  unwanted style patterns. Furthermore, we employ a learnable VGG~\cite{zhang2024artbank} to replace RDSMB to retrain StyMam. This substitution leads to a greater loss of content details and global features in the stylized images.
\section{Conclusion}
In this work, we propose StyMam which incorporates a dual-path strip scanning mechanism to capture fine-grained local textures and a channel-reweighted spatial attention module to preserve global dependencies. Compared with GAN-based approaches using CNNs or Transformers, StyMam achieves a more effective balance between local detail and global structural coherence. In contrast to SD-based methods, StyMam better preserves content structures while enabling faster inference. Extensive qualitative and quantitative evaluations demonstrate that StyMam outperforms state-of-the-art methods in both stylization quality and efficiency.
\bibliographystyle{IEEEbib}
\bibliography{strings,refs}

\end{document}